\documentclass[a4paper, 10pt, conference]{ieeeconf}      

\IEEEoverridecommandlockouts                              
\usepackage[utf8]{inputenc}
\usepackage[T1]{fontenc}

\usepackage{amsmath} 
\usepackage{amssymb}  
\usepackage{multirow}
\usepackage{makecell}
\usepackage{rotating}

\usepackage{booktabs}
\usepackage{cite}
\usepackage[textsize=tiny,final]{changes}

\definechangesauthor[color=orange]{DA}
\definechangesauthor[color=blue]{IG}

\title{\LARGE \bf
	Hybrid Soft Robots Incorporating Soft and Stiff Elements
}

\author{Dimuthu~D. K.~Arachchige and  Isuru~S.~Godage
	\thanks{$\!\!\!\!\!\!$Authors are with the Robotics and Medical Engineering (RoME) Laboratory, School of Computing, DePaul University, Chicago, IL 60604, USA.\,
		\vspace{1mm}
		\newline
		Corresponding author: {\tt\small DARACHCH@depaul.edu}\,
		\vspace{1mm}
		\newline 
		This work is supported in part by the National Science Foundation (NSF) Grants IIS-1718755, IIS-2008797, and IIS-2048142.
		\vspace{1mm}
		\newline
		This paper has been accepted to 2022 IEEE 5th International Conference on Soft Robotics (RoboSoft).
	}
}

\begin{document}
	
	\maketitle
	\thispagestyle{empty}
	\pagestyle{empty}
	
	\begin{abstract}
		Soft robots are inherently compliant and have a strong potential to realize human-friendly and safe 
		robots. 
%
		Despite 
		continued research highlighting
		the potential of soft robots, they remain 
		largely confined to laboratory settings.
		In this work, inspired by spider monkeys' tails, we propose
		a 
		hybrid soft robot (HSR) design. We detail the design objectives and methodology to improve controllable stiffness 
		range and achieve independent stiffness and shape control. 
		We extend the curve parametric 
		approach 
		to 
		We experimentally demonstrate
		that the proposed HSR has about 100\% stiffness range increase than a previous soft robot design with identical physical dimensions.
		In addition, we empirically map HSR's bending shape-pressure-stiffness and present an application example -- a soft robotic gripper -- to demonstrate the decoupled nature of stiffness and shape variations. 
		Experimental results show
		that proposed HSR can be successfully used in applications where independent stiffness and shape control is desired.
	\end{abstract}
	
	
	\section{Introduction\label{sec:Introduction}}
	Soft and continuum robots are inherently compliant structures 
	that undergo smooth and continuous structural deformation to form complex ``organic'' shapes. Prior work has demonstrated the potential of soft robots for adaptive whole arm grasping \cite{li2013autonomous}, obstacle avoidance and progressive planning \cite{li2016progressive}, grasping in cluttered space \cite{mcmahan2005design}, navigation in obstructive and unstructured environments \cite{meng2021anticipatory}, human-friendly interaction \cite{amaya2021evaluation}, and locomotion \cite{arachchige2021soft} just as few examples. Despite 
	continued research demonstrating 
	their immense potential, they are 
	largely confined to laboratory settings.
	This is mainly because of the lack of structural strength thereof necessary to engage in practical applications such as object manipulation or locomotion (move payload)
	while supporting their own weight. 
	
	%
	Soft robots are often actuated by pneumatic pressure, tendons, and smart materials\cite{schmitt2018soft}.
	An impressive number of prototypes that employ
	pneumatic muscle actuators (PMAs) have been proposed over the years \cite{walker2020soft}. PMAs are popular due to ease of customization, a wide operational pressure, and a high power-to-weight ratio \cite{caldwell1995control}. 
	The PMAs (typically 3) can be bundled together to construct bending robotic units -- termed sections \cite{trivedi2008geometrically}. Complex soft robots such as multisection manipulators or legged robots are then fabricated by combining many such bending units (sections).
	In addition to actuation, macro-scale PMA-powered soft 
	robots rely on PMAs 
	for the robot's structural integrity \cite{godage2015modal}. 
%
	The PMAs stiffness is proportional to the supply pressure. Thus, a soft robot constructed from multiple PMAs exhibits variable stiffness proportional to the mean pressure of PMAs. However, the achievable stiffness in PMAs solely through pressure increase is limited. This can lead to undesirable unpredictable (buckling behavior) and unstable (twisting that may result in permanent structural change) during operation.
	In addition, PMAs undergo length change proportional to the applied pressure. Consequently, the stiffness is coupled to robot shape. This means that it is not possible to change the robot stiffness without affecting the shape. This could be a problem in applications where adaptive stiffness control during taskspace trajectory tracking.

	
	
	\begin{figure}[t]
		\centering
		\includegraphics[width=0.9\linewidth]{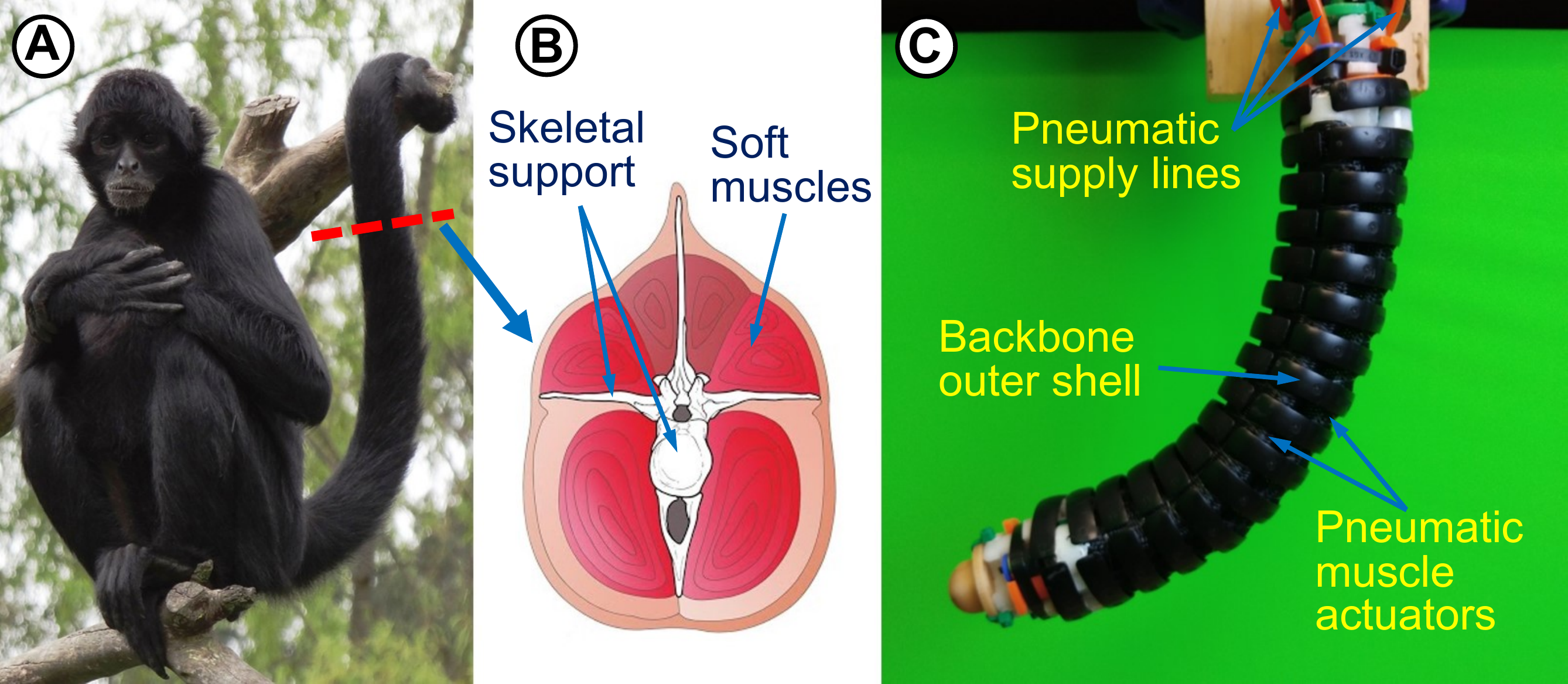}
		\caption{Bioinspiration -- (A) Spider monkey, (B) Tail’s muscular lining with skeletal support. (C) Proposed HSR prototype with bending.}
		\label{fig:Fig1_IntroductionImage}
	\end{figure}
	
	The 
	decoupled 
	stiffness and pose control allows soft
	robots to easily adapt to task demands in-situ without affecting the taskspace trajectories. However, prevalent continuum manipulators are made of sections that extend (or contract) depends on the operational mode of PMAs and are therefore subjected to length variation during operation.  Consequently, they lack the capability to independently control arm shape \cite{giannaccini2018novel}, and stiffness \cite{allaix2018stiffnessAp}. 
	Hence, a new line of thinking is warranted 
	for 
	potentially 
	generating technologies to bring soft robotics to practice while providing higher structural strength and better stiffness regulation to fulfill meaningful tasks without betraying compliant operation. 
	
	The usefulness of such features becomes evident when we consider a versatile biological example, such as the tail of spider monkeys (Fig. \ref{fig:Fig1_IntroductionImage}-A). Their muscular arrangement -- controlling the deformation of the skeletal structure underneath (Fig. \ref{fig:Fig1_IntroductionImage}-B) -- has embedded unique mechanical properties that differ from conventional soft robots. Muscles (and tendons) have spring-like properties, with inherent stiffness and damping. Thus, they can transform between modes (combinations of shape and stiffness) seamlessly. For instance, the tail can act as a manipulator (grasping tree branches), a support structure (standing upright), and a counterbalancing appendage (during jumping and climbing). They achieve these impressive transformations while still being ``soft'' to the touch because of the tail's muscular lining (Fig. \ref{fig:Fig1_IntroductionImage}-B). 
%
	Based on this  biological example, we propose a PMA powered hybrid soft 
	robot (HSR -- Fig. \ref{fig:Fig1_IntroductionImage}-C) with a rigid-linked, highly articulable but inextensible (i.e., constant-length) backbone. The primary motivations of combining soft elements (which facilitate smooth and continuous structural deformation while being compliant and human-friendly) with stiff elements (which provide structural strength to supporting high payload manipulation tasks) are, 1) to increase the variable stiffness range to adapt to environmental changes and operational needs, and 2) to decouple stiffness and shape change for achieving better motion control. 
	%
	%
	%

	\begin{figure}[t]
		\centering
		\includegraphics[width=1\linewidth]{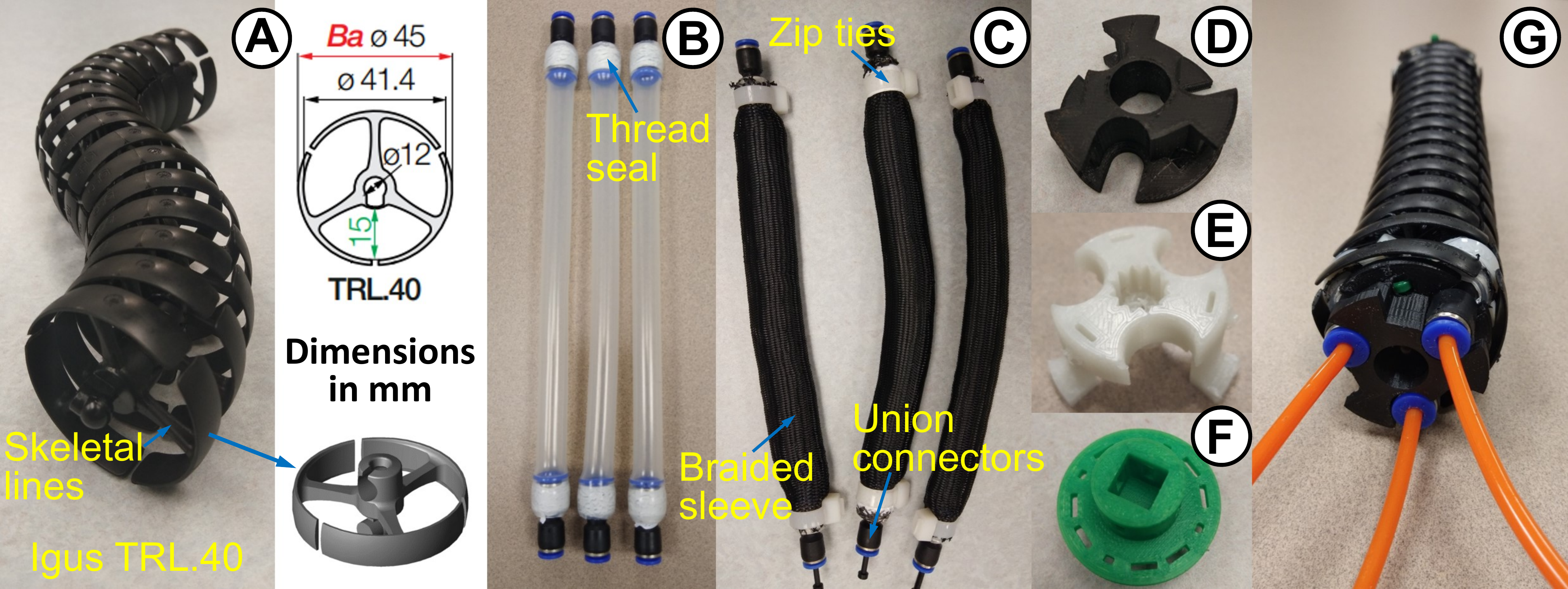}
		\caption{HSR design components -- (A) Igus Triflex R-TRL.40 dress pack, (B) Silicon tubes, (C) PMAs, (D) End caps, (E) Intermediate joint, (F) Upper hip joint. (G) Fabricated HSR prototype.}
		\label{fig:Fig2_PartDesign}
	\end{figure}
	
	\section{Hybrid Soft Robot (HSR) Design}\label{sec:design}
	
	The proposed HSR -- shown in Fig. \ref{fig:Fig1_IntroductionImage}-C -- has two main elements, namely the backbone and PMAs, that are used to produce the decoupled stiffness and deformation control. 
	\subsection{Highly-Articulable Backbone} \label{subsec:backbone}
	
	 
	 
	The backbone unit used in the design is a readily available dress pack (Igus Inc, 
	-- part no: Triflex R-TRL.40) designed for
	cable guide applications (Fig. \ref{fig:Fig2_PartDesign}-A). In the dress pack, 
	high-tensile-strength plastic segments
	are serially connected via ball-and-socket joints (allowing free rotation about the local $X$ and $Y$ axes) to form a rigid-link kinematic chain. The
	individual segments
	are easily assembled/disassembled allowing
	to customize the length of the backbone (i.e., section). We determined the backbone's length be $16~cm$ based on its bending ability to form a subtended angle of $180^\circ$ in any bending plane (Fig. \ref{fig:Fig2_PartDesign}-A).
	
	\subsection{Pneumatic Muscle Actuators (PMAs)}
	
	Custom-made Mckibben type PMAs are utilized to actuate the proposed HSR \cite{caldwell1993braided}. We used 
	commercially available Silicone tubes, pneumatic union connectors, braided sleeves, and heavy-duty zip ties to fabricate 
	PMAs \cite{godage2012pneumatic}. 
	We opted for the extending-mode PMAs
	for their leaner physical profile 
	as opposed to the contracting mode ones which would require a higher radius-to-length ratio.
	In addition, extending mode PMAs have higher normalized length variation (up to $50\%$ compared to about 35\% for contracting ones). Further, we desire comparably high operational pressures to generate 
	increased torques to achieve high bending deformation and stiffness range. To that end, we selected a silicone tube with $11~mm$ of inner diameter and $2~mm$ of wall thickness (Fig. \ref{fig:Fig2_PartDesign}-B). 
	We decided on the tube's thickness 
	based on the ability to safely operate 
	in 
	a pressure range (up to $5~bars$) with
	an acceptable 
	deadzone (pressure to overcome the transient radial expansion). 
	The length of a PMA is chosen as $150~mm$ based on the length and backbone characteristics such that the serial joints of the backbone can bend up to $180^\circ$ in any bending plane. To limit radial expansion of PMAs during operation, 
	the diameters of 
	constraining Nylon mesh 
	and Silicone tubes have to closely match. 
	Also, it increases the PMA efficiency 
	as most of the air pressure is used for axial extension. Consequently, this 
	further helps to reduce the PMA dead zone. To meet these design requirements, we experimentally selected high-strength Nylon braided mesh with diameters $10~mm$ (minimum) and $18~mm$ (maximum). We used 
	$4~mm$ internal diameter union connectors to connect external pneumatic pressure lines
	to PMAs. 
	Heavey duty
	zip ties are used to secure silicon tubes, mesh to the other end of the
	union connectors (Fig. \ref{fig:Fig2_PartDesign}-C.
	Fabricated PMAs shown in Fig. \ref{fig:Fig2_PartDesign}-C can extend by up to 50\%, and withstand $700~kPa$ with a $90~kPa$ pressure dead zone.
	
	\subsection{Hybrid Soft Robot (HSR) Assembly} \label{sub:integration}
	
	\begin{figure}[tb]
		\centering
		\includegraphics[width=0.85\linewidth]{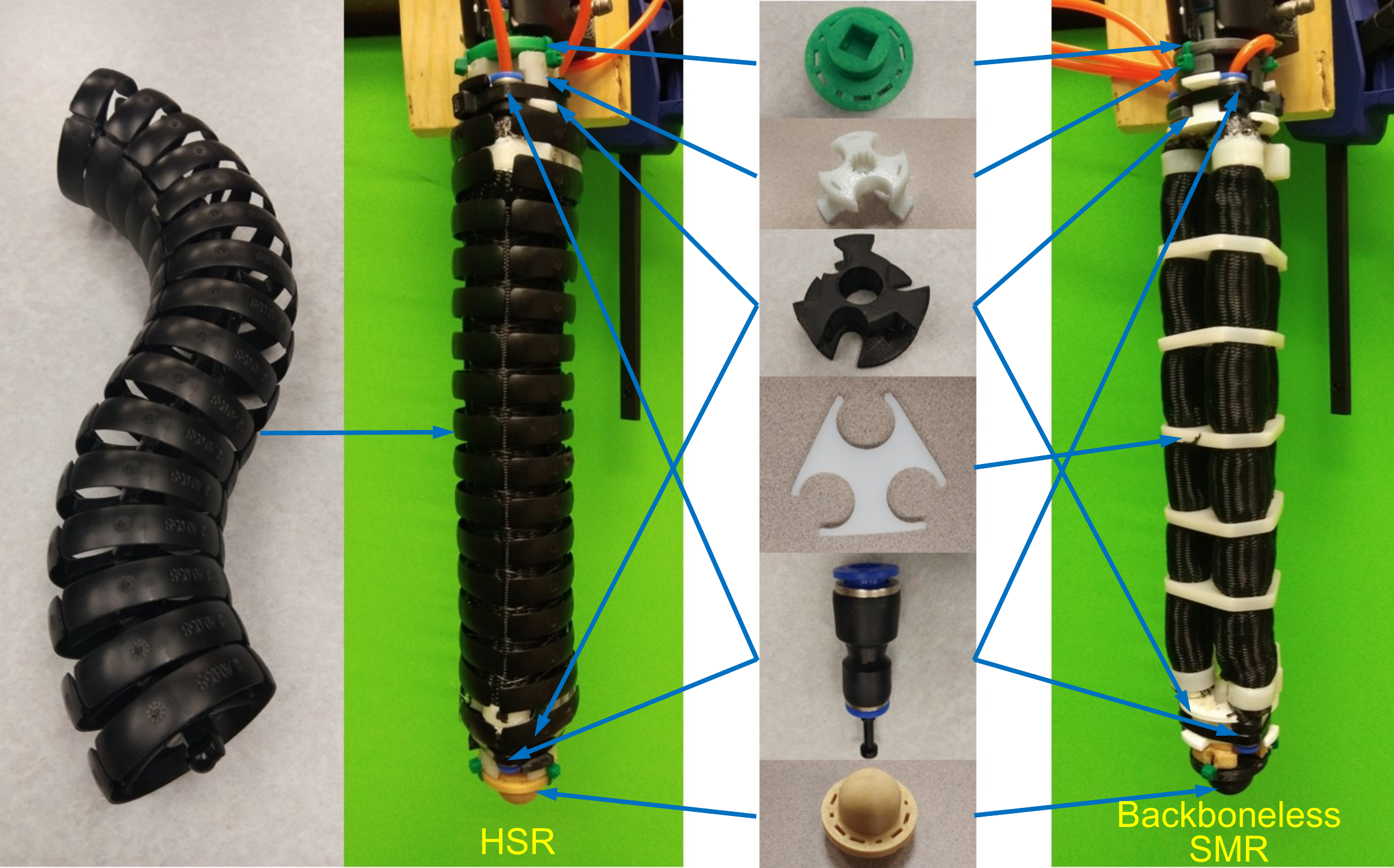}
		\caption{HSR and its counterpart with their design elements.}
		\label{fig:Fig3_UnitDesign}
	\end{figure}
	
	Fig. \ref{fig:Fig2_PartDesign}-G shows the finished HSR prototype. Therein, we arranged 3 PMAs within the radially symmetric cavities (or grooves) 
	along the length of the backbone structure (Fig. \ref{fig:Fig2_PartDesign}-A). We designed and 3D-printed several parts (Fig. \ref{fig:Fig2_PartDesign}-D, E, and F) to 
	integrate PMAs to the backbone.
	We used two end caps (Fig. \ref{fig:Fig2_PartDesign}-D) 
	to axially 
	secure the PMAs in place. 
	High-strength fasteners 
	then securely anchor the PMAs to the end caps
	at either end of the backbone.
	In this arrangement, the backbone constraints the PMA length change during operation which results in a spatially antagonistic PMA configuration for an increased stiffness range. 
	To prevent buckling and ensure uniform bending deformation during operation, we securely wrapped PMAs in parallel to the backbone
	using a fishing wire. 
	The resulting fixed-length HSR has a uniform construction and exhibits omnidirectional and circular arc bending. We employed 3D-printed joints
	shown in 
	Fig. \ref{fig:Fig2_PartDesign}-E and F 
	to connect HSRs 
	to the
	actuation base or each other. 
	Apart from the 
	fixed-length 
	HSR robot with backbone, to compare operable stiffness ranges as detailed in Sec. \ref{sub:expStiffnessRange}, we fabricated an identical variable-length soft 
	robot 
	prototype without a backbone (Fig. \ref{fig:Fig3_UnitDesign}). We used identical PMAs in both prototype designs. Fig. \ref{fig:Fig3_UnitDesign} provides an overview of used design elements in each prototype.
	
	\begin{figure}[tb]
		\centering
		\includegraphics[width=0.9\linewidth]{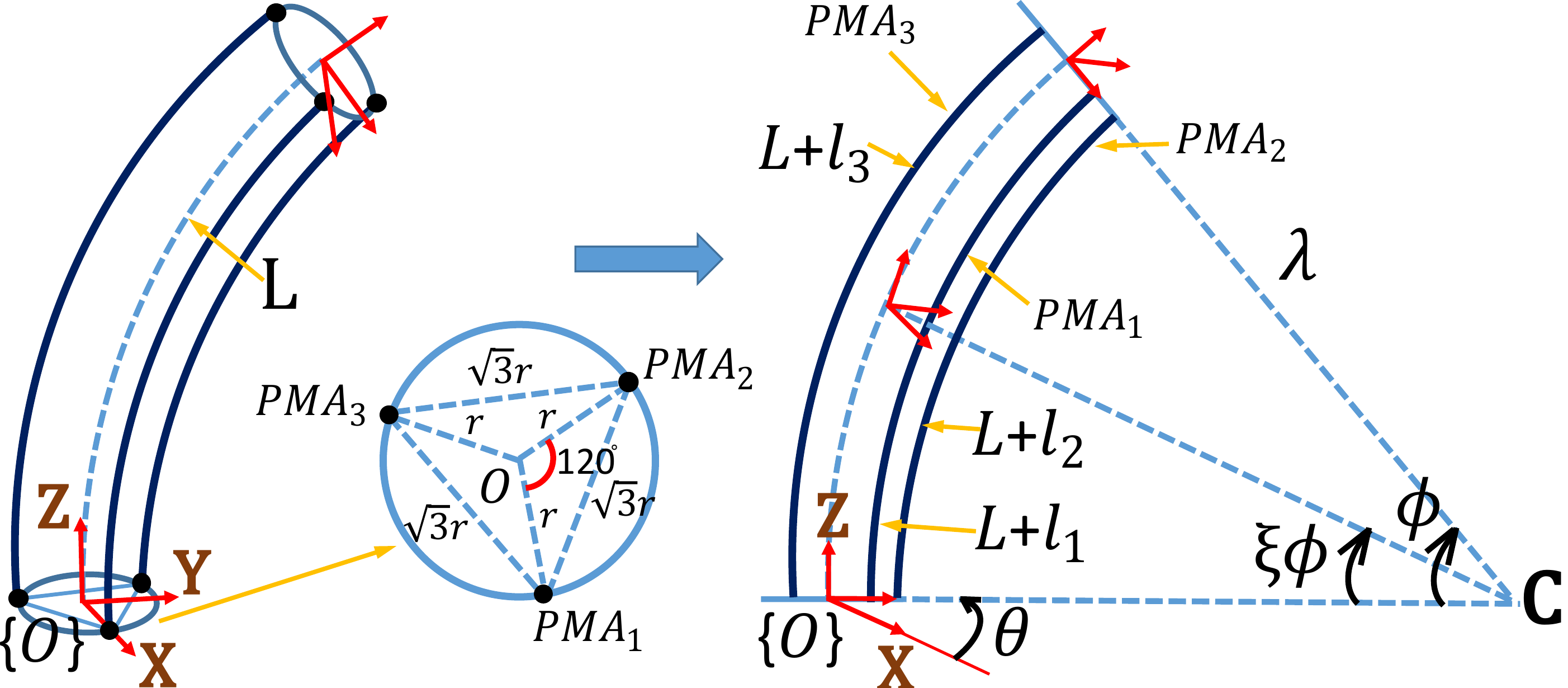}
		\caption{Schematic of the proposed HSR showing actuator arrangement.}
		\label{fig:Fig4_SchematicDiagram}
	\end{figure}
	
	\begin{figure}[tb]
		\centering
		\includegraphics[width=0.9\linewidth]{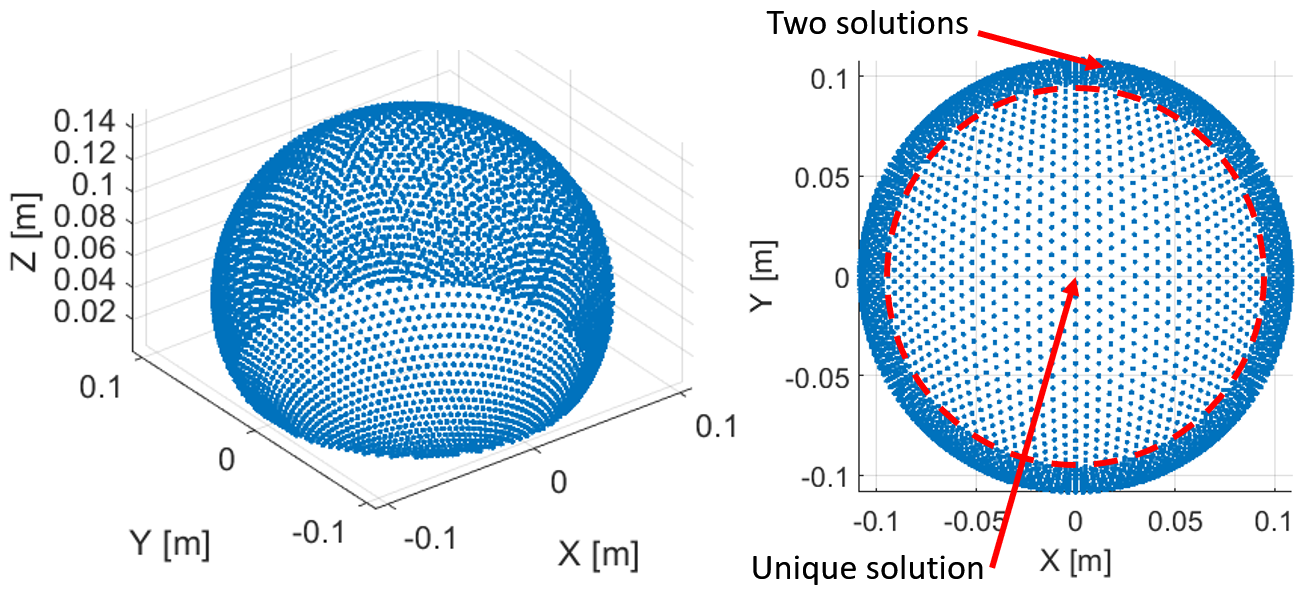}
		\caption{Spatial and top-down views showing the workspace symmetry.}
		\label{fig:Fig5_Taskspace}
	\end{figure}	
	
	\section{System Model}\label{sec:models}
	
		\begin{figure}[t]
		\centering
		\includegraphics[width=0.8\linewidth]{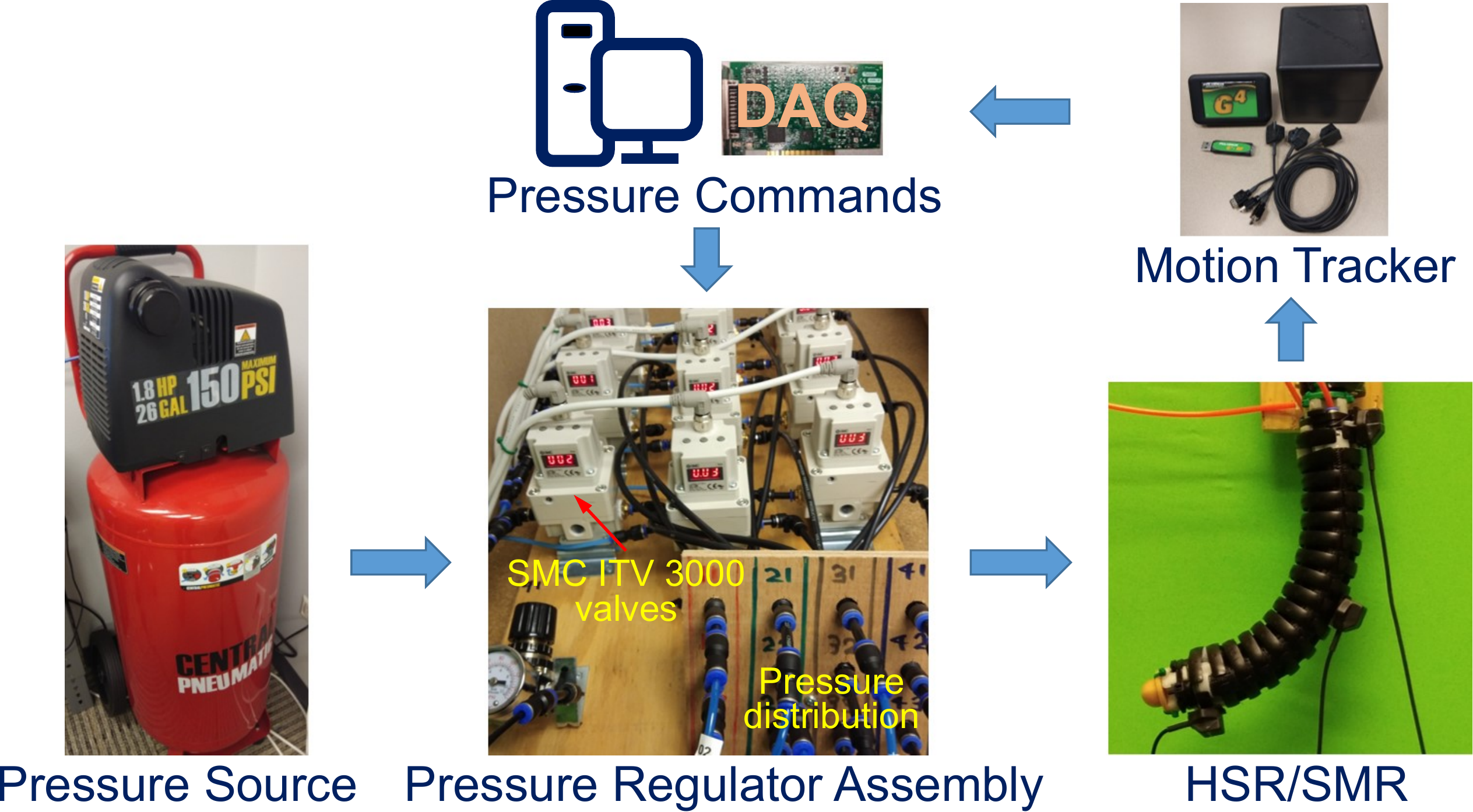}
		\caption{Experimental setups of the hybrid soft robot (HSR) and soft robots without backbone.}
		\label{fig:Fig6_ExperimentalSetup}
	\end{figure}
	\begin{figure*}[t]
		\centering
		\includegraphics[width=0.8\linewidth]{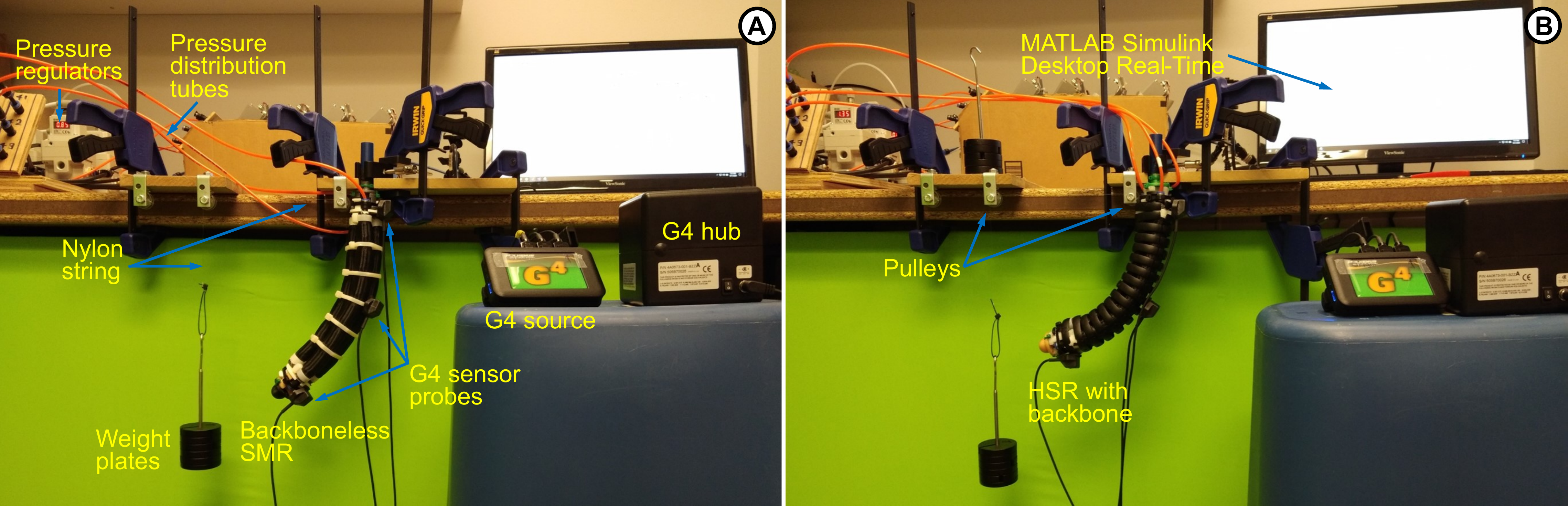}
		\caption{Experimental setups to obtain bending stiffness in, (A) Soft robot (without backbone), (B) Proposed HSR (with backbone).}
		\label{fig:Fig7_StiffnessTesting}
	\end{figure*}

	
	The kinematic modeling of the HSR is fundamentally similar to that of the variable-length continuum robots. However, the presence of length constraint introduces noteworthy characteristics that render it different in many respects including forward and inverse kinematic solutions. 

	Consider the schematic of the HSR shown in Fig. \ref{fig:Fig4_SchematicDiagram}. 
%
	Let the initial length of each actuator is $L$ whose change is in ${l_{i}} \in {\rm {\mathbb{R}}}$ where ${l_{i:\min }} \le {l_{i}}(t) \le {l_{i:\max }}$ for $i \in \{ 1,2,3\}$; $i$ and $t$ denote the actuator number and time, respectively. Therefore, ${L_{i}}(t) = L + {l_{i}}(t)$ calculates the actuator length at any time and the vector of joint variables of the HSR is defined as $\textbf{\textit{q}} = \left[ {{l_{1}}(t),\,{l_{2}}(t),\,{l_{3}}(t)} \right]^T$.
%
%
	We define the 
	HSR coordinate frame, $\{O\}$
	at the geometric center of the base (Fig. \ref{fig:Fig4_SchematicDiagram})
	with a PMA, whose length change is denoted by $l_1$, coinciding with the $+X$ axis at a radius $r$ from the origin.
	The PMAs denoted by $l_2$ and $l_3$ are situated in the counterclockwise direction at $\frac{2\pi}{3}$ and $\frac{4\pi}{3}$ angle offset from the $+X$, respectively. 
	When PMAs are actuated, forces resulting from elongation act on the endplates. 
	The force imbalance, due to a pressure differential in PMAs, 
	creates a net torque at the HSR tip,
	causing the robot
	to bend. 
	Given the uniform and radially symmetric construction, the HSR 
	neutral axis deforms 
	approximately in a circular arc shape. 
	(a reasonable assumption, as noted in \cite{godage2016dynamics}). 
	We parameterize the circular arc using 
	the
	subtended angle, 
	$\phi \in \left[0,\pi\right]$, and the angle to the bending plane with respect to the $+X$ axis, $\theta \in \left[0,2\pi \right)$. From the backbone length, $L$, we can derive the radius of 
	the circular arc as $\frac{L}{\phi}$. Employing arc geometry, we can relate the 
	PMA lengths to the 
	arc parameters as
	\vspace{1mm}
	\begin{align}
		L+l_{i} & =\left\{ \frac{L}{\phi }-r\cos\left(\frac{2\pi}{3}\left(i-1\right)-\theta \right)\right\} \phi \nonumber \\
		l_{i}& = -r\phi \cos\left(\frac{2\pi}{3}\left(i-1\right)-\theta \right)    	    	\label{eq:l2cp}
	\end{align}

	Note that, 
	due to the inextensible nature of HSR, the sum of PMA length changes 
	should add up to zero, {i.e., $\sum_{\forall i} l_{i} = 0$ from \eqref{eq:l2cp}} resulting from the kinematic constraint $l_{1}=-\left(l_{2}+l_{3}\right)$. 
	This implies that one DoF, out of 
	three, is redundant and the 
	HSR kinematics 
	can be derived by using
	two DoF. 
	The loss of DoF due to the
	kinematic constraint 
	can be used to 
	derive a reduced-order kinematic model as reported by \cite{deng2019near}. 
	We will leverage the
	redundant actuation DoF to control the stiffness and achieve independent stiffness and shape control. We use \eqref{eq:l2cp} to derive 
	arc parameters in terms of the jointspace variables as
	%
	
	\begin{subequations}
		\begin{align}
			\phi & =\frac{2\sqrt{\sum_{i=1}^{3}\left(l_{i}^{2}-l_{i}l_{\!\!\!\!\mod\left(i,3\right)+1}\right)}}{3r}\label{eq:cp_phi}\\
			\theta & =\arctan\left\{ \sqrt{3}\left(l_{3}-l_{2}\right),l_{2}+l_{3}-2l_{1}\right\} \label{eq:cp_theta}
		\end{align}
		\label{eq:cp2l}
	\end{subequations}
	Similar to \cite{godage2015modal}, the curve parameters are then used to derive the homogeneous transformation matrix for the HSR as
	\begin{align}
		\mathbf{T}\left(\boldsymbol{q},\xi\right) & =\mathbf{R}_{Z}\left(\theta\right)\mathbf{P}_{X}\left(\frac{L}{\phi}\right)\mathbf{R}_{Y}\left(\phi\right)\mathbf{P}_{X}\left(-\frac{L}{\phi}\right)\mathbf{R}_{Z}\left(-\theta\right)\nonumber \\
		& =\left[\begin{array}{cc}
			\mathbf{R}\left(\boldsymbol{q},\xi\right) & \mathbf{p}\left(\boldsymbol{q},\xi\right)\\
			0 & 1
		\end{array}\right]\label{eq:htm}
	\end{align}
	where $\textbf{R}_Z$ and $\textbf{R}_Y$ are homogeneous rotation matrices about $+Z$ and $+Y$ and $\textbf{P}_X$, is the homogeneous translation matrix along $+X$. $\textbf{R}$ and $\textbf{p}$ are the homogeneous rotation and position matrices, respectively. The scalar $\xi \in \left[0,1\right]$ denotes the points along the neutral axis with $\xi=0$ coinciding with the origin of the robot coordinate frame and $\xi=1$ pointing to the tip. Readers are referred to \cite{godage2015modal} for more details regarding the derivation.
	%
	
	Fig. \ref{fig:Fig5_Taskspace} shows the 
	HSR tip taskspace 
	generated using the kinematic model in \eqref{eq:htm}. 
	It is worth reporting that, in comparison to 
	designs without backbones (i.e., variable-length designs), the HSR taskspace is symmetric about the +Z axis of the robot coordinate system. 
	Whereas the taskspace of robot designs without backbones
	is tri-symmetric about the +Z axis. 
	In addition, the taskspace is essentially a thin shell (without volumetric coverage). In contrast, designs without backbones have a nonzero volume as a result of extension (or contraction) \cite{godage2015modal}. 
	The loss of a DoF also 
	results in a one-to-one mapping between the jointspace and curve parameters (configuration space). 
	This feature lets one perform the computations using curve parameters and then 
	derive 
	jointspace solutions. This approach is not feasible with variable-length robots as the curve parameters describing the arc radius and subtended angle are coupled \cite{godage2015modal}.

	\section{Experimental Validation}\label{sec:results}
	
	
	\subsection{Experimental Setups}\label{sub:exp_setup}
	A block diagram of the experimental setup is shown in Fig. \ref{fig:Fig6_ExperimentalSetup}. The air compressor provides a constant $8~bar$, input air pressure. 
	Each PMA pressure is
	controlled by an SMC ITV3000 (Orange Coast Pneumatics, Inc. USA) digital proportional pressure regulator. 
	Input pressure commands, generated by a MATLAB Simulink Desktop Real-time model, communicated to pressure regulators via an
	NI PC-6704 data acquisition card. 
	To capture the HSR taskspace movement, three wireless 6-DoF trackers are used 
	(Polhemus G4 wireless -- Polhemus, Inc. USA). 
	We mounted trackers at either end -- base and tip -- as well as at the mid-point of the HSR (Refer to Fig. \ref{fig:Fig7_StiffnessTesting}- A \& B) and recorded the tracking data at $100~Hz$. 
	The complete experimental setups for both HSR and the variable-length counterpart (i.e., soft robot without backbone) are shown in Fig. \ref{fig:Fig7_StiffnessTesting}. 	
	
	\begin{table}[tb]
		\footnotesize
		\setlength{\tabcolsep}{2pt}
		\centering
		\caption{Stiffness variation in soft robot without backbone and 
			HSR}
		\label{table:bendingstiffness}
		\begin{tabular}{cc|ccccccc}
	\begin{sideways}\end{sideways} & \multicolumn{1}{c}{} &  & \multicolumn{5}{c}{Bending stiffness [$Nm/ rad$]} &  \\
	\multicolumn{1}{l}{\begin{sideways}\end{sideways}} & \multicolumn{1}{l}{} & \multicolumn{7}{c}{without backbone} \\
	\multirow{7}{*}{\rotcell{$P_1~[bar]$}} & 3.0 & x & x & x & x & x & x & {1.39} \\
	& 2.5 & x & x & x & x & x & {1.22} & {1.32} \\
	& 2.0 & x & x & x & x & {0.94} & {1.07} & {1.12} \\
	& 1.5 & x & x & x & {0.62} & {0.71} & {0.86} & {0.92} \\
	& 1.0 & x & x & {0.49} & {0.57} & {0.63} & {0.70} & {0.73} \\
	& 0.5 & x & {0.42} & {0.46} & {0.51} & {0.59} & {0.62} & {0.62} \\
	& 0.0 & {0.39} & {0.40} & {0.43} & {0.48} & {0.53} & {0.58} & {0.61} \\ 
	\cline{3-9}
	\begin{sideways}\end{sideways} & \multicolumn{1}{c}{} & 0.0 & 0.5 & 1.0 & 1.5 & 2.0 & 2.5 & 3.0 \\
	\begin{sideways}\end{sideways} & \multicolumn{1}{c}{} &  & \multicolumn{5}{c}{$P_2~[bar]$} &  \\
	\multicolumn{1}{l}{} & \multicolumn{1}{l}{} & \multicolumn{1}{l}{} & \multicolumn{5}{l}{} & \multicolumn{1}{l}{} \\
	\multicolumn{1}{l}{} & \multicolumn{1}{l}{} & \multicolumn{1}{l}{} & \multicolumn{5}{c}{Bending stiffness [$Nm/ rad$]} & \multicolumn{1}{l}{} \\
	\multicolumn{1}{l}{} & \multicolumn{1}{l}{} & \multicolumn{1}{l}{} & \multicolumn{5}{c}{with backbone} & \multicolumn{1}{l}{} \\
	\multirow{7}{*}{\rotcell{$P_1~[bar]$}} & 3.0 & x & x & x & x & x & x & {3.21} \\
	& 2.5 & x & x & x & x & x & {2.60} & {2.84} \\
	& 2.0 & x & x & x & x & {1.76} & {1.96} & {2.40} \\
	& 1.5 & x & x & x & {0.97} & {1.26} & {1.63} & {1.87} \\
	& 1.0 & x & x & {0.70} & {0.82} & {0.97} & {1.21} & {1.45} \\
	& 0.5 & x & {0.56} & {0.68} & {0.75} & {0.82} & {0.93} & {1.24} \\
	& 0.0 & {0.52} & {0.54} & {0.62} & {0.70} & {0.79} & {0.90} & {1.12} \\ 
	\cline{3-9}
	\multicolumn{1}{l}{} & \multicolumn{1}{l}{} & 0.0 & 0.5 & 1.0 & 1.5 & 2.0 & 2.5 & 3.0 \\
	\multicolumn{1}{l}{} & \multicolumn{1}{l}{} & \multicolumn{1}{l}{} & \multicolumn{5}{c}{$P_2~[bar]$}
	\end{tabular}
	\end{table}
	\normalsize	
	
	\subsection{Validate Stiffness Control Range Improvement}\label{sub:expStiffnessRange}
	
\begin{figure}[tb]
		\centering
		\includegraphics[width=0.85\linewidth]{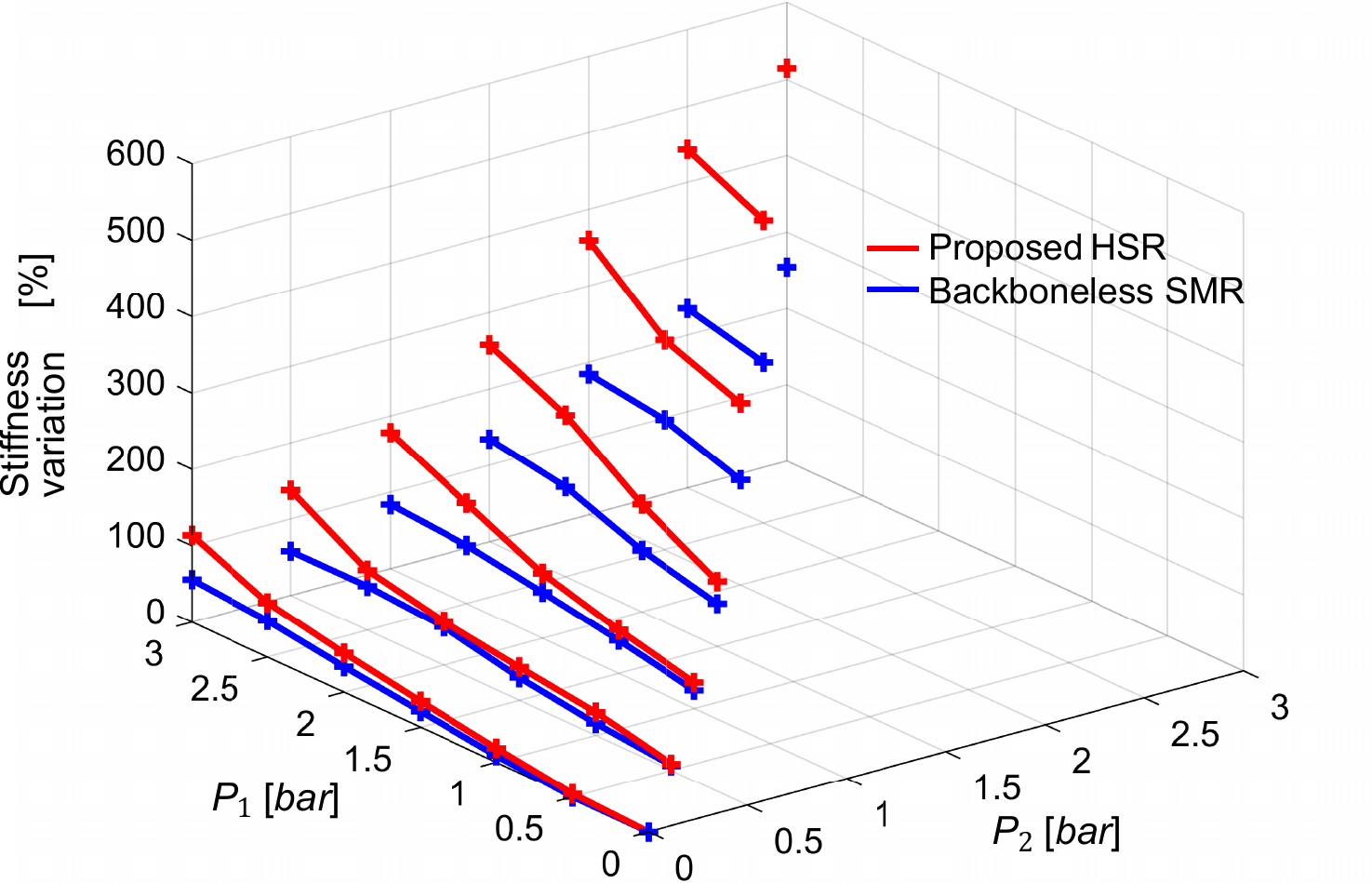}
		\caption{Percentage change of bending stiffness in two soft robot designs.}
		\label{fig:Fig8_BendingStiffness}
	\end{figure}

	This section assesses the impact of backbone integration on the improvement of controllable stiffness range with respect to the robot designs with and without backbone, shown in Fig. \ref{fig:Fig7_StiffnessTesting}-A and B respectively. 
	To limit the torsional deformation during operation, we constrain
	the bending of 
	both designs to a plane. 
	We achieved this by
	simultaneously supplying the same pressure commands to
	two PMAs. 
	This results in planar bending deformation as shown in Fig. \ref{fig:Fig8_BendingStiffness}. 
	The other pressure input counteracts the bending and, due to the antagonistic PMA arrangement and fixed-length constraint, controls the stiffness. 
	We use an experimental 
	approach to estimate the stiffness. Thus, we can use the same arrangement and limit outside influences associated with setup changes. 
	Further, 
	as 
	we 
	replicate 
	the same test in both soft robot designs with identical actuator arrangements, we can extrapolate and generalize the results for omnidirectional bending across both designs. The supply pressure combinations that were applied in these tests are shown in Table \ref{table:bendingstiffness}. 
	To comply with the limitations of the experimental setup, 
	the pressure combinations are chosen such that the bending is unidirectional, as shown in Fig. \ref{fig:Fig8_BendingStiffness}. We achieve this behavior by setting a higher or equal pressure value to $P_2$ pressure component (corresponds to the pressure of two simultaneously actuated PMAs). 
	In this experiment,
	we use pressure combinations outlined in Table \ref{table:bendingstiffness} for both designs. 
	Upon applying each 
	pressure 
	combination, we utilized a pulley arrangement to 
	provide 
	a bending torque perturbation.
	Note that the torque perturbation is normal to the neutral axis of the prototype (Fig. \ref{fig:Fig7_StiffnessTesting}). 
	We recorded the 
	change in bending angle ($\Delta \phi$)
	due to the torque perturbation ($\Delta \tau$). 
	Note that, the motion tracker 
	system 
	provides taskspace data
	in terms of positions and orientations. 
	We used the 
	kinematic model in Sec. \ref{sec:models} to 
	derive the arc parameters, $\phi$, and $\theta$ from the taskspace data.
	We calculated the bending stiffness ($K$) using $K=\frac{\Delta \tau}{\Delta \phi}$. We repeated the same procedure with similar pressure inputs on the variable-length soft robot (i.e., no backbone) design.  
	
	\begin{figure}[tb]
		\centering
		\includegraphics[width=1\linewidth]{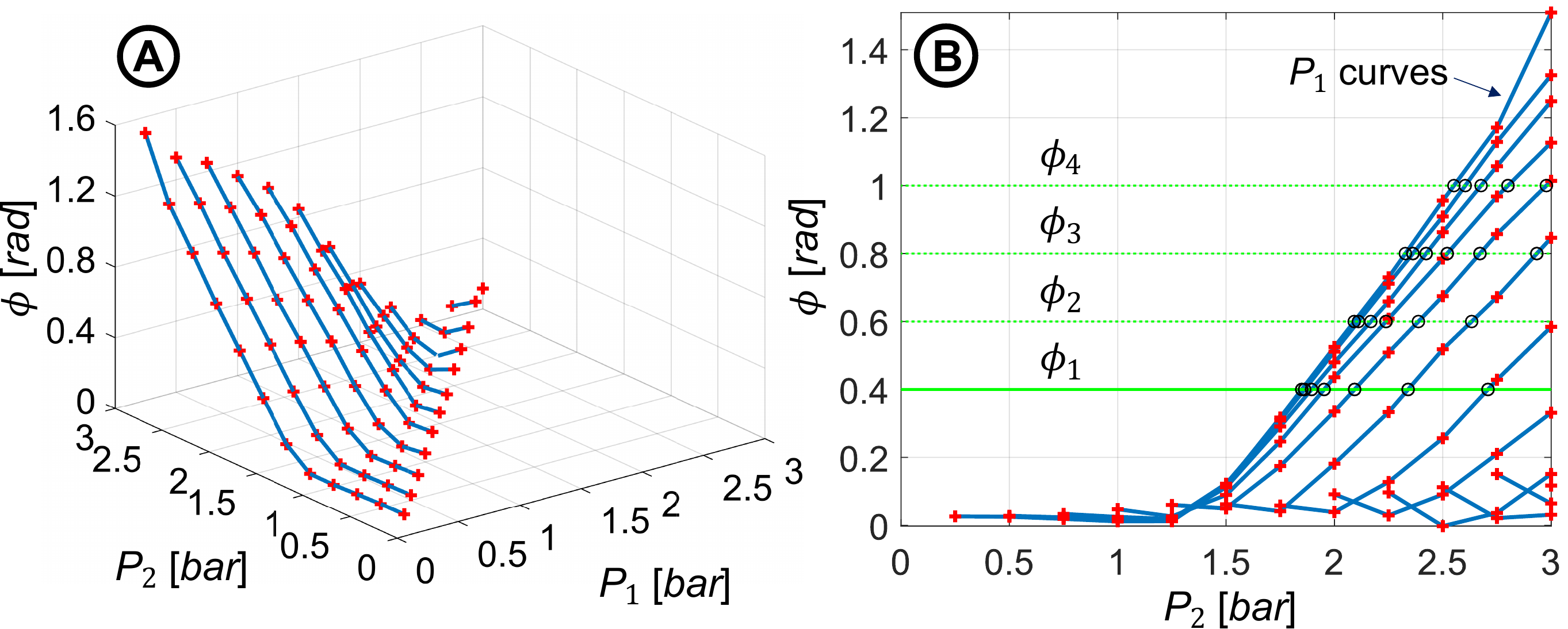}
		\caption{(A) Shape-pressure mapping of HSR, (B) Extracting pressure combinations for identified planner bending shapes.}
		\label{fig:Fig9_P1P2Phi}
	\end{figure}
	
	Stiffness variation computed through perturbation tests
	are presented in Table \ref{table:bendingstiffness}-Top for the backboneless soft robot and in Table \ref{table:bendingstiffness}-Bottom for the proposed HSR. 
	The percentage bending stiffness variation with respect to the natural bending stiffness -- at $P_1=0~bar$ and $P_2=0~bar$ --  is shown in Fig. \ref{fig:Fig8_BendingStiffness}. The percentage increase of the bending stiffness in the soft robot without a backbone is $\frac{(1.39-0.39)}{0.39} \times  100\ \% = 256.41\ \%$. Similarly, the percentage bending stiffness increase in the proposed HSR is $\frac{(3.21-0.52)}{0.52} \times  100\ \% = 517.31\ \%$. The results show that the bending stiffness increase in the proposed HSR is about 100\% higher than that of the variable-length soft robot. 
	Note that both designs have comparable natural stiffness (when applied pressures are 0). This result is expected because both designs use identical PMAs and physical arrangements sans the backbone. Thus, the results conclusively demonstrate that the backbone integration significantly enhances the operational bending stiffness range without betraying bending or compliance capability. With the added benefit of on-demand stiffening capability, it supports tasks requiring structural strength such as variable payload handling.
	
	\begin{table}[tb]
		\footnotesize
		\setlength{\tabcolsep}{2pt}
		\centering
		\caption{Shape-pressure-stiffness mapping of HSR.}
		\label{table:mappingtable}
			\begin{tabular}{|c|c|c|c|} 
				\hline
				\multirow{2}{*}{\begin{tabular}[c]{@{}c@{}} \textbf{HSR shape}\\$\phi$~$[rad]$\\ \end{tabular}} & \multicolumn{2}{c|}{\begin{tabular}[c]{@{}c@{}}\textbf{Pressure}\\\textbf{combinations} \end{tabular}} & \begin{tabular}[c]{@{}c@{}}\textbf{Bending }\\\textbf{stiffness}\end{tabular} \\ 
				\cline{2-3}
				& ${P_1}~[bar]$  & ${P_2}~[bar]$  & \multicolumn{1}{l|}{$[Nm/rad]$} \\ 
				\hline
				\multirow{4}{*}{0.4} & 0.50 & 1.86 & 0.63 \\ 
				\cline{2-4}
				& 0.75 & 1.90 & 0.81 \\ 
				\cline{2-4}
				& 1.00 & 1.96 & 1.11 \\ 
				\cline{2-4}
				& 1.25 & 2.09 & 1.32 \\ 
				\hline
				\multirow{4}{*}{0.6} & 0.50 & 2.11 & 0.71 \\ 
				\cline{2-4}
				& 0.75 & 2.17 & 0.85 \\ 
				\cline{2-4}
				& 1.00 & 2.24 & 1.40 \\ 
				\cline{2-4}
				& 1.25 & 2.39 & 1.71 \\ 
				\hline
				\multirow{4}{*}{0.8} & 0.50 & 2.36 & 0.86 \\ 
				\cline{2-4}
				& 0.75 & 2.42 & 1.42 \\ 
				\cline{2-4}
				& 1.00 & 2.52 & 1.90 \\ 
				\cline{2-4}
				& 1.25 & 2.67 & 2.18 \\ 
				\hline
				\multirow{4}{*}{1.0} & 0.50 & 2.60 & 1.56 \\ 
				\cline{2-4}
				& 0.75 & 2.68 & 1.98 \\ 
				\cline{2-4}
				& 1.00 & 2.80 & 2.33 \\ 
				\cline{2-4}
				& 1.25 & 2.98 & 2.58 \\
				\hline
			\end{tabular}
		\end{table}
		\normalsize	
		
		\subsection{Validate Decoupled Stiffness and Deformation Control}\label{sub:decoupleStiffnessShape}
		
		We empirically evaluated the independent stiffness and shape controllability of the proposed HSR using the experimental setup in Fig. \ref{fig:Fig7_StiffnessTesting}-B.  First, we recorded the shape variation ($\phi$) against a range of actuation pressure combinations ($P_1, P_2$) using the motion tracking
		system under no-load conditions. Note that, here we applied bidirectional bending similar to Section \ref{sub:expStiffnessRange}.  Fig. \ref{fig:Fig9_P1P2Phi}-A shows the recorded shape variation and Fig. \ref{fig:Fig9_P1P2Phi}-B shows the 2D plot of it. Marked bending shapes ($\phi_{1}-\phi_{4}$) in Fig. \ref{fig:Fig9_P1P2Phi}-B show that, along a particular bending plane (say $\phi_{1}$), we can extract several corresponding pressure combinations (highlighted in circles). Next, we applied a load profile and measured the corresponding bending stiffness (similar to Section \ref{sub:expStiffnessRange}) for four extracted pressure combinations on $\phi_{1}$ bending plane. Subsequently, we repeated the procedure for other bending planes ($\phi_{2}$, $\phi_{3}$, \&, $\phi_{4}$) and Table \ref{table:mappingtable} shows extracted pressure combinations and recorded bending stiffness values. The table data reveals that different stiffness values are recorded for the same bending angle under different pressure combinations indicating independent stiffness and shape controllability of the proposed HSR.
		%
		%
		\subsection{Application Example: A  
			Soft Robotic Gripper}\label{sub:gripper}
		
\begin{figure}[tb]
			\centering
			\includegraphics[width=1\linewidth]{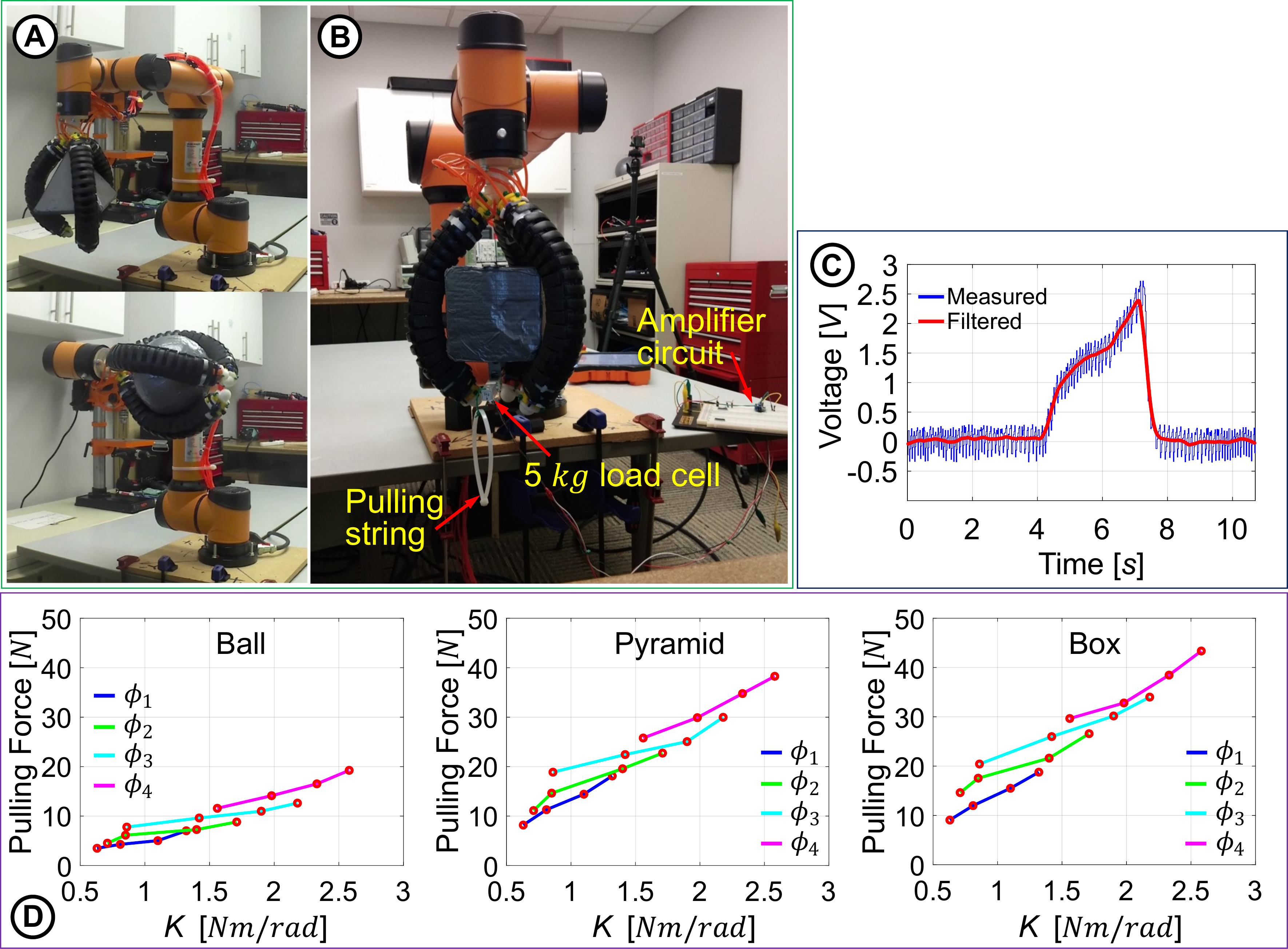}
			\caption{(A) Proposed variable stiffness gripper with pyramid and ball objects, (B) Force test experiment with box object, (C) Pulling force reading, (D) Failure forces of three objects.}
			\label{fig:Fig10_ForceTestmaximumvoltage}
		\end{figure}

		We combined three HSR units to 
		fabricate a tri-fingered soft robotic gripper shown in Fig. \ref{fig:Fig10_ForceTestmaximumvoltage}-A \cite{arachchige2021novel}. 
		We conducted the experiment shown in Fig. \ref{fig:Fig10_ForceTestmaximumvoltage}-B to quantify the effect of decoupled stiffness and shape control
		for 
		improving 
		grip quality 
		without exerting pressure on objectes.
		We 
		quantify
		grip quality 
		utilizing the external force perturbation needed to release an object from the grip. In that respect, higher forces indicates better grasp.
		Three objects with geometrically varying surfaces but with the same smoothness (pyramid, ball, and box -- approximately $100~g$ weight) 
		were gripped by applying pressure combinations recorded in Table \ref{table:mappingtable} to HSR fingers. And in each actuation, an external pulling force is applied to the object via an attached cable until the grip fails. We coupled a $5~kg$ load cell 
		to record failure forces. Fig. \ref{fig:Fig10_ForceTestmaximumvoltage}-C shows a typical force sensor output 
		during the
		test. Therein, we uniformly increase the force 
		while recording the force sensor data
		until the grip fails. Then we apply a 50-sample moving average to 
		filter the noise in force sensor data 
		and measure the peak value (failure force).
		Fig. \ref{fig:Fig10_ForceTestmaximumvoltage}-D presents maximum failure forces against each object's stiffness variation. The plots show that under the same finger shape (same $\phi$), the grip failure force increases with the bending stiffness for each object. Moreover, higher bending (i.e., firmer grip $\rightarrow$ $\phi_{1}<\phi_{2}<\phi_{3}<\phi_{4}$) has recorded 
		higher failure forces 
		for each object. Further, ball pulling and box pulling have recorded relatively the smallest and largest failure forces, respectively. This is due to the fact that ball and box have the relatively lowest and the highest irregular surfaces among the three object shapes. The force test results reveal that the proposed HSR units are useful in applications where independent stiffness and shape control is desired.
		%

		\section{Conclusions}\label{sec:conclusions}
		This paper proposed a novel HSR that combines rigid and soft elements in the construction. The HSR design is inspired by the spider monkey tails, where the muscles lining the inextensible vertebra control the shape and stiffness. We hypothesized that an articulable and inextensible backbone with PMAs could mimic the same biological characteristics. We extended the curve parametric approach to account for fixed-length design and derive the 
		HSR kinematic model. We experimentally compared the stiffness variation of the proposed HSR with an identical 
		soft robot without a backbone and showed that the new HSR has an approximately 100\% higher stiffness increase. Next, we conducted a pressure-shape-stiffness mapping experiment and showed that the proposed HSR can independently control the stiffness and shape. Subsequently, we presented an application example -- a soft robotic gripper constructed from the proposed HSR units -- to further validate the HSR's independent stiffness and shape controllability. The results showed that the new HSR can still operate the same way without significantly affecting taskspace performance or the natural compliance with the added benefit of a high stiffness range. This stiffness range opens up possibilities in field applications where adaptive stiffness is required, such as being human-friendly and supporting high payload capabilities beyond current soft robotic designs.	
		
		\bibliographystyle{IEEEtran}
		\bibliography{refs}
		
	\end{document}